\documentclass[sigconf]{acmart}
\usepackage{multirow}
\usepackage{graphicx}
\usepackage{subfigure}
\AtBeginDocument{%
  \providecommand\BibTeX{{%
    \normalfont B\kern-0.5em{\scshape i\kern-0.25em b}\kern-0.8em\TeX}}}

\acmConference{DLP-KDD 2020}{August 24, 2020}{San Diego, California, USA} 
\acmBooktitle{Woodstock '18: ACM Symposium on Neural Gaze Detection,
  June 03--05, 2018, Woodstock, NY}
\acmPrice{15.00}
\acmISBN{978-1-4503-XXXX-X/18/06}
\usepackage[utf8]{inputenc} 
\usepackage[T1]{fontenc}    
\usepackage{hyperref}       
\usepackage{url}            
\usepackage{booktabs}       
\usepackage{amsfonts}       
\usepackage{nicefrac}       
\usepackage{microtype}      
\usepackage{graphicx}
\usepackage{algorithm}
\usepackage{algorithmic}
\usepackage{xcolor}
\usepackage{multirow}

\begin{document}


\title{Automated Model Selection for Time-Series Anomaly Detection}
\author{Yuanxiang Ying, Juanyong Duan, Chunlei Wang, Yujing Wang, Congrui Huang, Bixiong Xu}
    \affiliation{ 
      \institution{Microsoft}
      \state{Beijing}
      \country{China}
    }
    \email{{yuyi,juaduan,chuwan,yujwang,conhua,bix}@microsoft.com}

\renewcommand{\shortauthors}{Yuanxiang and Juanyong, et al.}
\begin{abstract}
	Time-series anomaly detection is a popular topic in both academia and industrial fields. Many companies need to monitor thousands of temporal signals for their applications and services and require instant feedback and alerts for potential incidents in time. The task is challenging because of the complex characteristics of time-series, which are messy, stochastic, and often without proper labels. This prohibits training supervised models because of lack of labels and a single model hardly fits different time series. In this paper, we propose a solution to address these issues. We present an automated model selection framework to automatically find the most suitable detection model with proper parameters for the incoming data. The model selection layer is extensible as it can be updated without too much effort when a new detector is available to the service. Finally, we incorporate a customized tuning algorithm to flexibly filter anomalies to meet customers' criteria. Experiments on real-world datasets show the effectiveness of our solution.
\end{abstract}



\maketitle

\section{Introduction}
Modern corporations need to monitor millions of temporal signals to make sure their business are working healthily. Any aberrant signal may indicate troubles which could cause significant revenue loss. In-time detection of such anomalies is necessary and could trigger prompt troubleshooting as soon as possible to avoid such loss. For example, business units of a company can rely on the detection system to alert any drops in sales so that they can come up with solutions in time. From the perspective of data science, time-series anomaly detection aims to discover unexpected events or rare items in data. The task requires to label any data point that is different from the majority of data points. For this purpose,an anomaly detector helps customers monitor the time-series continuously. 

\begin{figure}
	\centering
	\includegraphics[width=0.48 \textwidth]{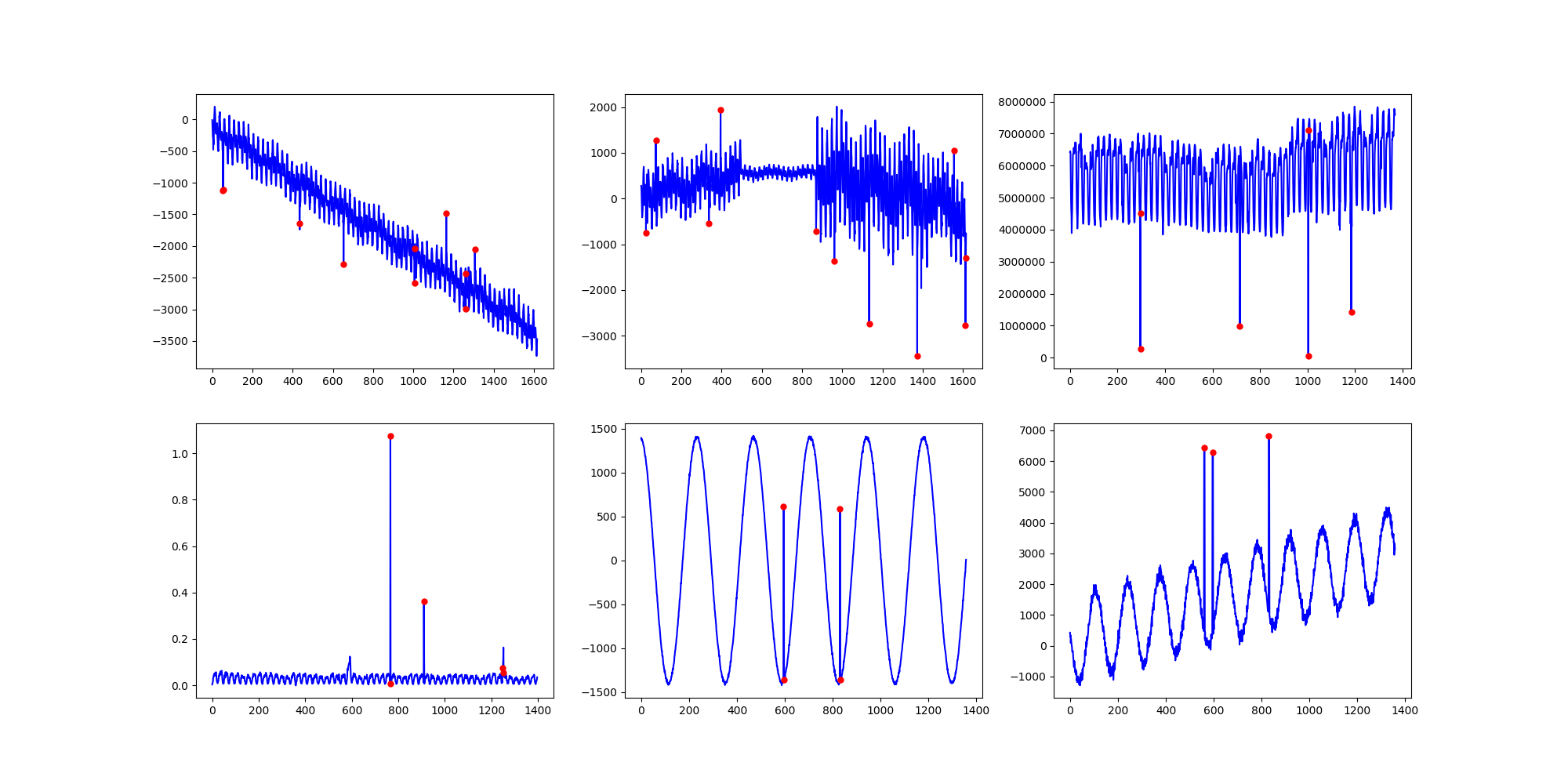}
	\caption{Patterns of Time-Series}
	\label{fig:patterns}
\end{figure}

Many challenges could be foreseen in designing such an industrial service. First, supervised models may be inefficient for this task due to lack of labels and continuous changing of data distribution. Since customers always have substantial time-series to be monitored, providing anomaly labels could be ineffective. The data distribution may change from time to time so the pretrained models may not fit newly coming data. Second, the patterns of time-series are very complex as shown in Figure~\ref{fig:patterns}. The anomaly detector is expected to work well on all kinds of time-series, which requires the model to have good generalization capability. Previous methods may only achieve good performance on a specifit type of data but fail on others. For example, Holt winters \cite{chatfield1978holt} can detect anomalies accurately on seasonal signals but performs poorly on non-seasonal data. Third, customers have different tolerance on anomalies. Whether a data point is anomaly depends on customers' strategy and business concerns. Therefore, they need an intuitive and effective way to customize the anomaly detection results.

\begin{figure*}[t]
    \centering
    \subfigure[DataFeed]{
		\includegraphics[width=0.45\textwidth, height=3.5cm]{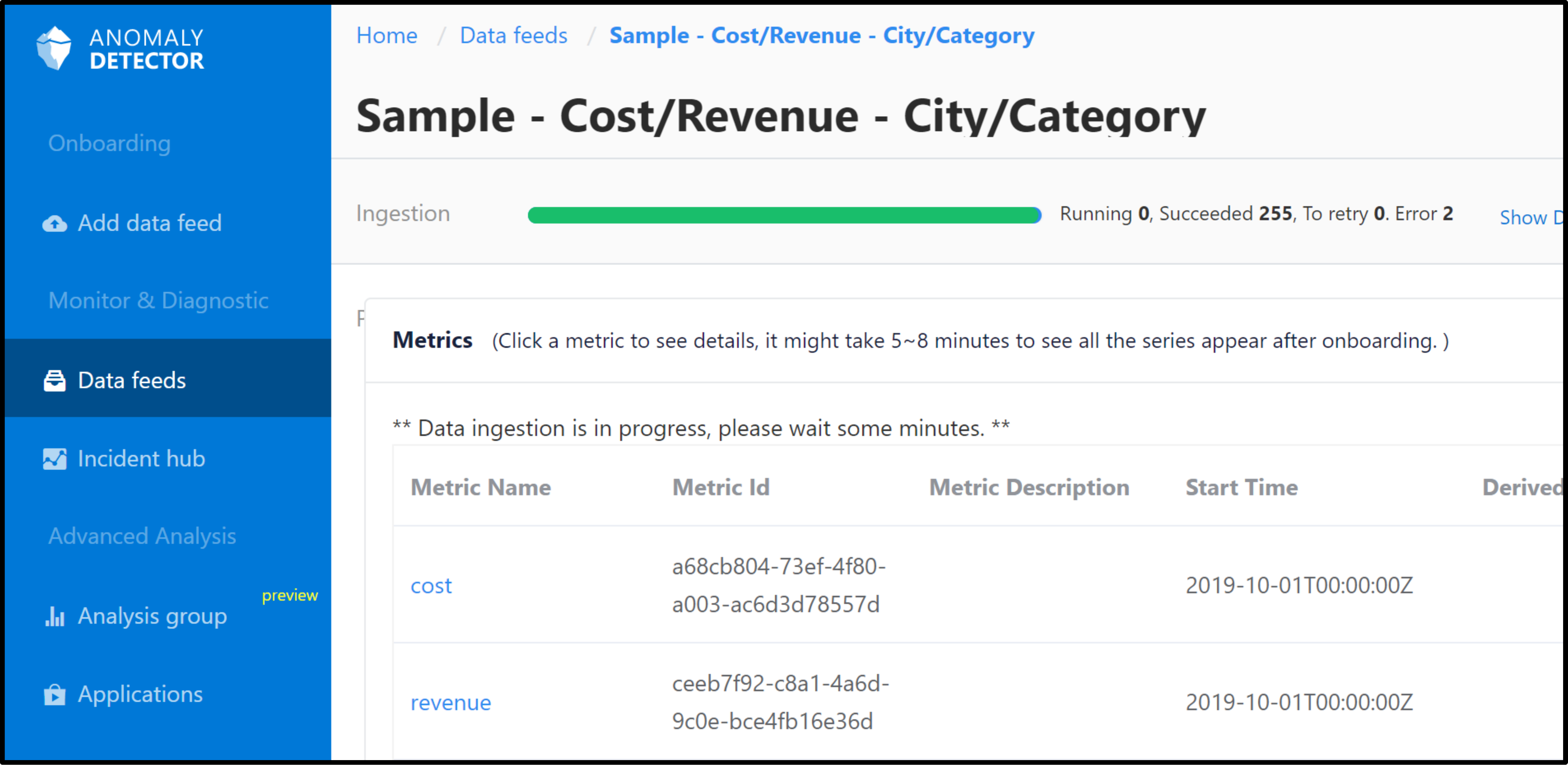}}
	 \subfigure[Detection Result]{
	    \includegraphics[width=0.45\textwidth, height=3.5cm]{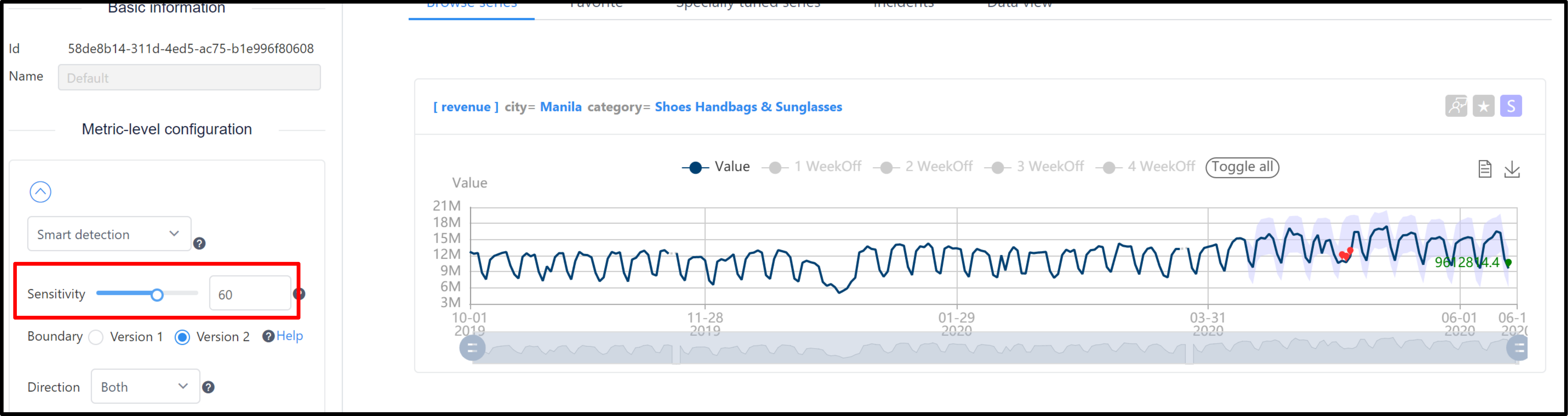}}
	 \caption{Application Overview}
	 \label{fig:sample}
\end{figure*}

At Microsoft, it is a common need to monitor business metrics and act quickly to address the issue if there is anything outside of the normal pattern. To tackle the problem, we build a scalable system with the ability to monitor minute-level time-series from various data sources. Automated diagnostic insights are provided to assist users to resolve their issues efficiently. The service has been used by more than 200 product teams within Microsoft, across M365, Bing and Azure, with more than 4 million time-series ingested and monitored continuously. As shown in Figure~\ref{fig:sample}, users create multi-dimension datafeed to collect their service metrics, including DevOps, Service Quality and User Activity. The monitor system will detect anomalies for each metrics and users can filter the detection result with a single parameter named \textit{Sensitivity}.

In this work, we present our solution on combining multiple time-series anomaly detection models. We design an automated model selection pipeline for time-series anomaly detection which selects the most suitable detection model based on the features of input time-series. All the candidate models are unsupervised which will be discussed in section~\ref{SR-CNN}. Our pipeline is also equipped with a parameter estimator which could compute the best hyper-parameters for the selected anomaly detector. Finally, we provide a tuning interface to allow customized interpretation of anomaly results. Experiments have shown the effectiveness of our design and the pipeline can be integrated into customers' own platforms and business units.
\section{Methodology}
\begin{figure}
	\centering
	\includegraphics[width=0.48 \textwidth]{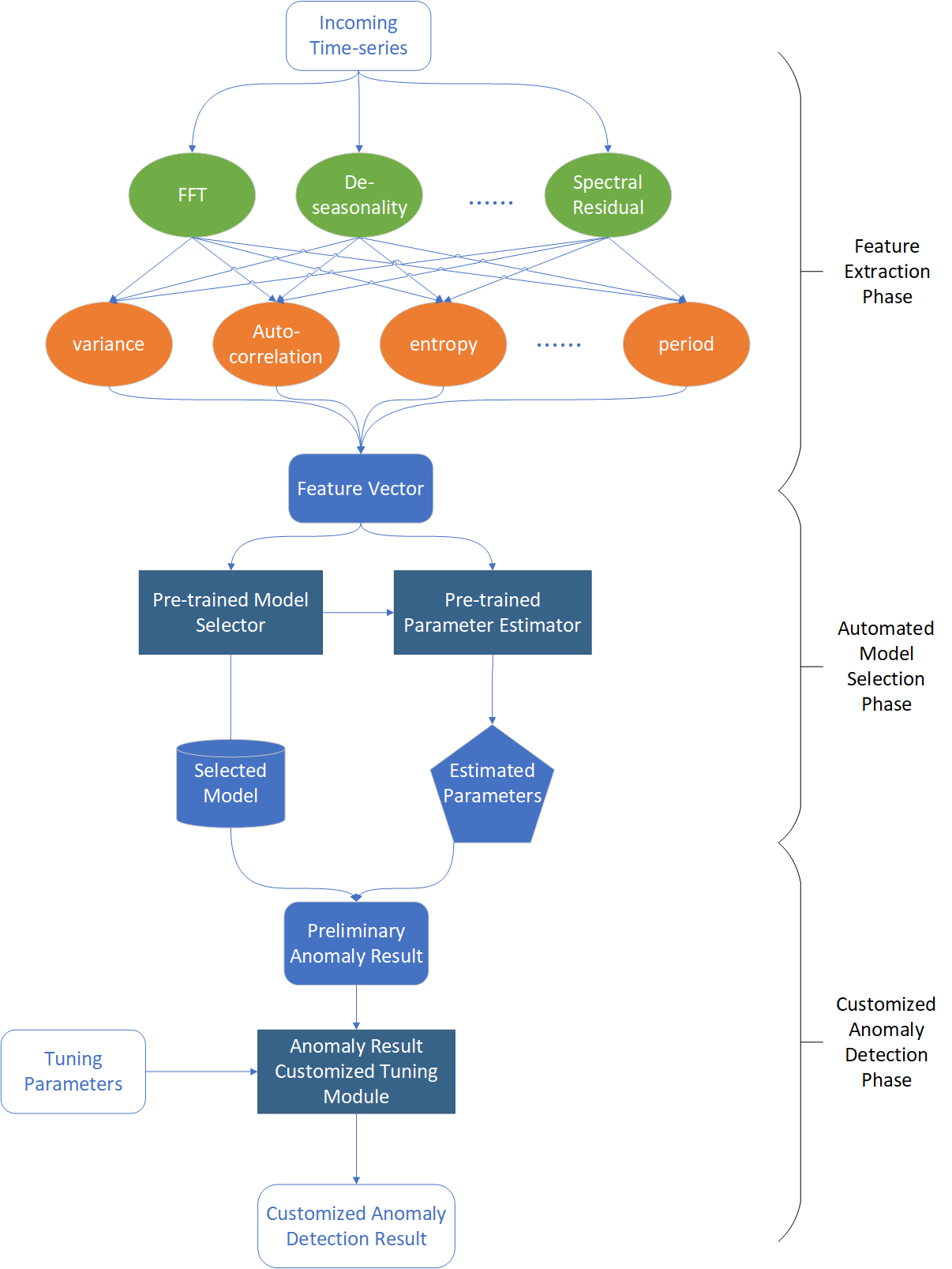}
	\caption{Pipeline}
	\label{fig:pipeline}
\end{figure}
In this section, we discuss about the key components in our anomaly detector. An overview of the whole pipeline is described in Section~\ref{overview}. Then we introduce how we leverage Spectral Residual in time-series anomaly detection in Section~\ref{SR-CNN}, methodology of building pre-trained anomaly detection model selector in Section~\ref{Model Selector} and the ability of customized tuning in Section~\ref{Customized Tuning}.
\subsection{Overview}
\label{overview}
The whole pipeline of our service is presented in Figure~\ref{fig:pipeline}. The incoming series is first processed by a set of transformations and feature extractors. Then in the automated model selection phase, Model Selector takes the extracted features as input and outputs the anomaly detection model that best fits the input data. Each anomaly detection model is associated with a Parameter Estimator, which is used to compute related parameters. Next, our service uses the selected model and its corresponding parameters to detect anomalies of the input data and obtains a preliminary anomaly detection result. Lastly, tuning parameters are applied to obtain a customized anomaly detection result.

\textbf{Feature Extraction Phase}
To obtain richer feature representations, we apply a list of transformations on the input time series, namely, Spectral Residual, FFT, De-seasonality and so on and get a list of transformed series. Then we compute the common statistic metrics, like mean and variance, and a selected set of time-series features on the original time series, as well as on these transformed series. The final dimension of features is 82. The transformations and feature extractors can be extended based on needs. We use a fixed window size $\omega$ to segment time series to which simulates the real scenario that a limited length of historical points are seen when detecting anomaly on the latest point.

\textbf{Automated Model Selection Phase}
In this phase, the pre-trained model  selector and parameter estimator are invoked to select an anomaly detection model for the incoming series automatically. The model selector has been designed as a classification task. The goal of this classifier is to select a proper anomaly detection model based the extracted features. For each base anomaly detection model, there is an associated parameter estimator that predicts key parameters. We use a regression task to learn the fitted hyper-parameters for each kind of anomaly detection model. Details of these two modules will be discussed in section~\ref{Model Selector}.

\textbf{Customized Anomaly Detection Phase}
Once the anomaly detection model and parameters are known, a preliminary anomaly result can be obtained by invoking that model to perform detection on the input series. Then the result is refined by the Anomaly Result Customized Tuning module with user-defined tuning parameters. Finally, the customized anomaly detection result is represented to the users. Details of the customized anomaly detection result tuning algorithm is discussed in section~\ref{Customized Tuning}.
\subsection{Anomaly Detection Models}
\label{SR-CNN}
One of the main challenge in time-series anomaly detection is the various pattern of time-series. Our approach \textbf{Auto-Selector} aims to select the best anomaly detection model and its corresponding hyper-parameters for each series. On serving an industrial time-series anomaly detection service, we need to consider latency and generalization. Such limitations motivate us to select model candidates from statistical time-series anomaly detection models or unsupervised models since these models can achieve stable performance efficiently~\cite{sr-cnn}. In this work, \textbf{Auto-Selector} selects best model from three candidates as follow:
\begin{itemize}
    \item \textbf{SR}~\cite{sr-cnn} transforms time-series into the frequency domain, while anomalies are filtered with a certain threshold after the transformation. This threshold is an important parameter for SR and is included in the \textbf{Parameter Estimator} phase.
    \item \textbf{HBOS}~\cite{goldstein2012histogram} calculates the probability of being anomaly for each data point based on the histogram with a certain threshold of probability. The threshold used for probability filtering is considered in our \textbf{Parameter Estimator} phase.
    \item \textbf{S-H-ESD}~\cite{hochenbaum2017automatic} detects anomalies by Extreme Studentized Deviate test (ESD) after removing trend and seasonal information. In S-H-ESD, the max anomaly ratio has been used to control whether an individual point will be detected as an anomaly. Therefore, we consider the max anomaly ratio as an important hyper-parameter for S-H-ESD.
\end{itemize}
\subsection{Pre-trained Anomaly Detection Model Selector}
\label{Model Selector}
In real world applications the distribution of time-series is often stochastic and difficult to predict. A single anomaly detection model is hardly capable of predicting all time-series correctly. Meanwhile, the performance of an anomaly detection model relies heavily on suitable hyper-parameters. To address these issues, we propose a model selector that can select the most suitable anomaly detection model and its associated hyper-parameters for each time-series automatically. Specifically, the model selector has been learned offline and serve online in Anomaly Detector.

Model selector aims to find a proper detection model $m$ for each series $s$ from the limited model space $\mathcal{M}$. In order to train the model selector, we build model space with popular anomaly detection models. Although the scale of time series is tremendous, the detection models and series patterns are limited. Assuming we have anomaly labels of different series pattern, the model selector problem is actually a multi-classification problem. Therefore, we gather series with anomaly labels as a time-series knowledge base. Constantly enrich the knowledge base, we will get superior model selector. Presently, we combine a multi-classifier with a heuristic classifier to generate model selector which has been used in Anomaly Detector. The multi-classifier is trained with the time-series knowledge base and the heuristic classifier could ensure the detection quality once a series pattern hasn't been included in the knowledge base.

 Specifically, the algorithm~\ref{alg_featuralization} shows how we extract feature vector $\vec{U}$ for each series. Given a time-series $\vec{v}$, $n$ transformations $t_i(\vec{v}), i \in [1, n]$ are applied on $\vec{v}$ to obtain $n$ transformed series $\vec{w_i}, i \in [1, n]$. Then $k$ feature extractors are applied on each transformed series to generate $n$ feature vectors $\vec{u_i} \in \mathbb{R}^k$. Those features are then used to train the multi-classifier. With this classifier, a model with confidence score $\mathcal{C}$ can be inferred for each series. If the classifier cannot obtain confident result, a heuristic classifier will provide empirical result for the series.  
\begin{algorithm}
	\caption{Series Feature Extraction}
	\label{alg_featuralization}
	\begin{algorithmic}
		\REQUIRE $\vec{v}$, $t_1(\cdot)$, $t_2(\cdot)$, ..., $t_n(\cdot)$, $f_1(\cdot)$, $f_2(\cdot)$, ..., $f_k(\cdot)$
		\ENSURE $\vec{U}$
		\FOR {$i \gets 1$ to $n$}
		\STATE $\vec{w_i} \gets t_i(\vec{v})$
		\STATE $\vec{u_i} \gets (f_1(\vec{w_i}), f_2(\vec{w_i}), ..., f_k(\vec{w_i}))$
		\ENDFOR
		\STATE $\vec{U} \gets (\vec{u_1}, \vec{u_2}, ..., \vec{u_n})$
	\end{algorithmic}
\end{algorithm}



The mechanism of generating model selector makes it easier to update the selector so that Anomaly Detector can keep pace with cutting-edge anomaly detection techniques over time. Updating time-series knowledge base and bring in new anomaly detection models are convenient in this process. Concretely speaking, we build an offline pipeline to automate the iterative development process of model selector. This offline pipeline first passes through the time-series knowledge base to know the best anomaly detection model and associated parameters for each series. Then it trains model selector to learn these knowledge by updating the transformations, feature extractors, multi-classifiers and parameter estimators. If there are gains in the update, a set of gated test are run against the new model selector to check if the quality, efficiency and stability of the new model selector can reach the release standard.

\subsection{Customized Anomaly Detection Result Tuning}
\label{Customized Tuning}
Customers may have the requirement of intuitively tuning the results as the tolerance on anomalies varies in different real scenarios. To this end, we propose a method to customize the anomaly detection result with a single parameter $\alpha \in [0, 100]$. The larger $\alpha$ is, the more anomalies will be reported. Let $\vec{v}\in\mathbb{R}^n$ be the input time-series of length $n$ and we compute through our pipeline the initial anomaly detection result $\vec{a}\in\mathbb{R}^n$, where $a_i=1$ if $v_i$ is an anomaly otherwise 0 for $i=1,\cdots,n$. We then apply Algorithm \ref{alg_tuning} to adjust $\vec{a}$. First, $\vec{v}$ is decomposed into three components $\vec{g}$, $\vec{s}$ and $\vec{\epsilon}$ which represent the trend, the seasonality, and the loss component of $v_i$ respectively. We compute a delta unit $\vec{\mu}$ based on the trend component and estimate the final tolerant range $\vec{\delta}$ through a factor function, where $\vec{\delta}$ is the maximum loss each point can accept and the factor function is an exponentially decreasing function w.r.t. $\alpha$. When the fluctuation is small comparing with the magnitude of the normal values, $\alpha$ is set to be smaller than 50, otherwise it should be larger than 50. Finally, an anomaly point will be labeled as normal if the absolute value of its loss is within its maximum tolerable loss. To make the tuning process interactive, we have designed a UI as shown in Figure \ref{fig:sample}. The purple area around the series reflects the tolerant range under given $\alpha$. Users can change $\alpha$ and find the shape of the purple area changes gradually along with the anomaly detection results.

\begin{algorithm}
	\caption{Anomaly Detection Result Tuning}
	\label{alg_tuning}
	\begin{algorithmic}
		\REQUIRE $\vec{v}$, $\vec{a}$, $\alpha$
		\ENSURE $\vec{a'}$
		\STATE $\vec{g}, \vec{s}, \vec{\epsilon}  \gets \textbf{decompose}(\vec{v})$ \\
		\STATE $\vec{\mu} \gets 0.5 \cdot |\vec{g}| + 0.5 \cdot \sum_{j=1}^{n} |g_j|/ n$
		\STATE $\vec{\delta} \gets \textbf{factor}(\alpha) \cdot \vec{\mu}$
		\STATE $\vec{a'} \gets \vec{a} \cdot \mathbf{1}(\vec{\delta} > |\vec{\epsilon}|)$
	\end{algorithmic}
\end{algorithm}

\section{Experiments}
\subsection{Datasets and Metrics}

\paragraph{Datasets} We use an internal time-series dataset TSD that is collected in our production to evaluate our model. The statistics of this dataset is shown in Table~\ref{table:dataset}.  This dataset is randomly divided into two parts as the train set and test set by ratio $3: 1$. For each anomaly detector $\mathcal{M}_i$ we select a set of candidate parameters \{$p_1^i, p_2^i, \cdots, p_n^i$\}. Then for each pair $(\mathcal{M}_i, p_j^i)$, $j=1,\cdots,n$ we compute the F1 score of anomaly detection on each time series and find the best pair as the ground truth label for the model selector and the parameter estimator.

\paragraph{Metrics} We use F1 score, precision, and recall to evaluate the performance of our model for anomaly detection. In practice, anomalous observations usually form contiguous segments since they occur in a continuous manner. Following the evaluation strategy of~\cite{sr-cnn}, we mark the whole segment as positive sample if any observation in this segment is detected as an anomaly correctly. In real application, we expect our model discover anomalies promptly in order to take actions. Only if there is a point in an anomaly segmentation and the delay is no more than $k$ from the start of the segmentation, we mark the whole segmentation as successful detection. In our experiment, $k$ is set to $1$.

\begin{table}[h!]
	\caption{Dataset}
	\label{table:dataset}
		\begin{tabular}{p{5cm}c}\hline
			& TSD \\\hline
			Number of sequences & 1570  \\
			Dataset size & 308,822 \\
			Number of anomalies & 23,356 \\
			Anomaly ratio & 7.56\% \\\hline
		\end{tabular}%
\end{table}

\subsection{Experiment Settings}
In our experiments, besides SR~\cite{sr-cnn}, we select S-H-ESD~\cite{hochenbaum2017automatic} and HBOS~\cite{goldstein2012histogram} as our base anomaly detection models. For each anomaly detection model, a key hyper-parameter has been estimated in our experiments. In SR and HBOS, we estimate threshold that has been used to filter anomalies. In S-H-ESD, we estimate max anomaly ratio which has been used to control the detection tolerance. The window size of this experiment $\omega$ is 29. The model selector is implemented as an LightGBM classification model and we implement a parameter estimator for each anomaly detector as an LightGBM regression model. Once the model selector outputs the candidate model, we call its corresponding parameter estimator to compute the optimal parameters.

\begin{table}[h!]
	\caption{Quantitative Results}
	\label{table:quantitative_result}
	\begin{tabular}{c|ccc}
		\hline
		Model    & F1   & Precision  & Recall \\ \hline
		SR & 0.3446    & 0.2724    & 0.4688 \\ 
		HBOS & 0.4190    & 0.3753     & 0.4743 \\ 
		S-H-ESD & 0.4148 &  0.3595 & 0.4901 \\ 
		Auto-Selector & $\textbf{0.4738}$ & 0.4780 &  0.4692 \\ \hline
	\end{tabular}%
\end{table}

\begin{table}[h!]
	\caption{Quantitative Results on Best Parameter}
	\label{table:parameter_estimator_result}
	\begin{tabular}{c|ccc}
		\hline
		Model    & F1   & Precision  & Recall \\ \hline
		SR & 0.4719    & 0.5137    & 0.4364 \\
		HBOS & 0.4969    & 0.3910     & 0.6817 \\
		S-H-ESD & 0.4711    & 0.3651     & 0.6638 \\ 
		Auto-Selector & \textbf{0.5714}    & 0.4909     & 0.6837 \\
		\hline
	\end{tabular}%
\end{table}



\subsection{Results and Discussion}
Table~\ref{table:quantitative_result} compares Auto-Selector with base single anomaly detection model. Parameter of each single anomaly detection model has use its best parameter searched in the training set, specifically, threshold of SR is 2, threshold of HBOS is 0.99 and max anomaly ratio of S-H-ESD is 0.01. We could see Auto-Selector has improved $F1$ score significantly which indicates our approach has the ability to select best fitted model for each individual series.

In Table \ref{table:parameter_estimator_result}, we demonstrate $F1$ result with best parameter which means hyper-parameter has been searched on the test set instead of using the \textbf{Parameter Estimator} in Auto-Selector. Thus experiment aims to analyze the performance of \textbf{Model Selector}. Compared with \textbf{Model Selector}, \textbf{Parameter Estimator} aims to select best hyper-parameter in a broader search space so it should be difficult for this module to get satisfied results. We also select best fitted hyper-parameter on the test set for each single anomaly detection model. Obviously, the performance of each single anomaly detection model has been improved in different extent. With the best fitted parameter, Auto-Selector also improve its $F1$ score more than 20\%. The analysis result show that the trained model selector can provide fitted anomaly detection model for each series.
\section{Online Serving}
In this section, we first describe how this framework is integrated into our product in ~\ref{online_serving_stack}, then we report two cases that the Automated Model Selection help improve the anomaly detection quality in our product.
\subsection{Online Architecture}
\label{online_serving_stack}
\begin{figure}
	\centering
	\includegraphics[width=0.45\textwidth]{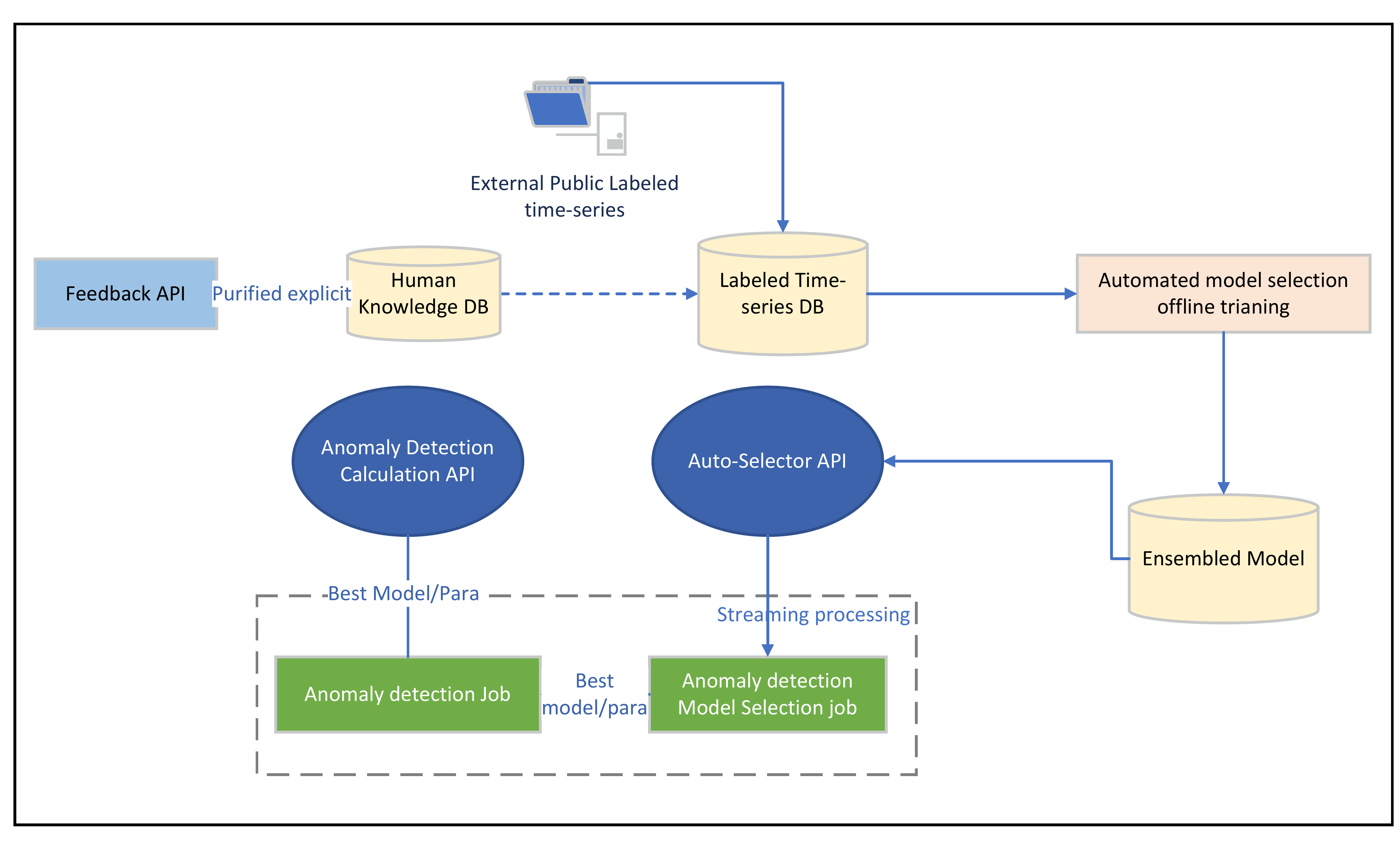}
	\caption{Online Architecture}
	\label{fig:serving_stack}
\end{figure}
We deploy the end-to-end framework into our online monitoring system as containers\footnote{https://kubernetes.io/docs/concepts/containers/} on AKS\footnote{https://azure.microsoft.com/en-us/services/kubernetes-service/}. Several main components have been described in Figure~\ref{fig:serving_stack}. Periodically, the offline training module will re-train the \textbf{Auto-Selector} to include updated labeled time-series and update the selection candidates of time-series anomaly detection models. In online serving, a model selection API will be used to serve the offline-trained \textbf{Auto-Selector} model to select the best model and its corresponding parameters for each individual series. The \textit{Anomaly detection Job} and \textit{Anomaly detection Model Selection Job} serve together to detect anomalies for each timestamp of the input metrics in streaming. The model selection job will trigger re-selection based on the anomaly rate and false alert rate of a time-series. The anomaly detection job will leverage the selected model to detect anomalies with the \textit{Anomaly Detection Calculation API}. Moreover, we design the feedback mechanism to collect labels from users. Those feedback are added into the labeled time-series conditionally.
\subsection{Case Study}
\label{case_study}
\begin{figure*}[t]
    \centering
    \subfigure[Detection results of Single Model]{
		\includegraphics[width=0.45\textwidth, height=3.5cm]{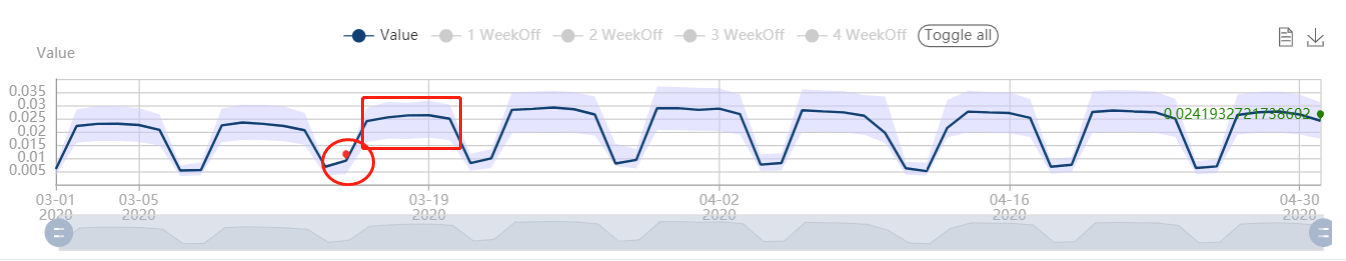}
		\label{case1-a}}
	 \subfigure[Detection results of Auto-Selector]{
	    \includegraphics[width=0.45\textwidth, height=3.5cm]{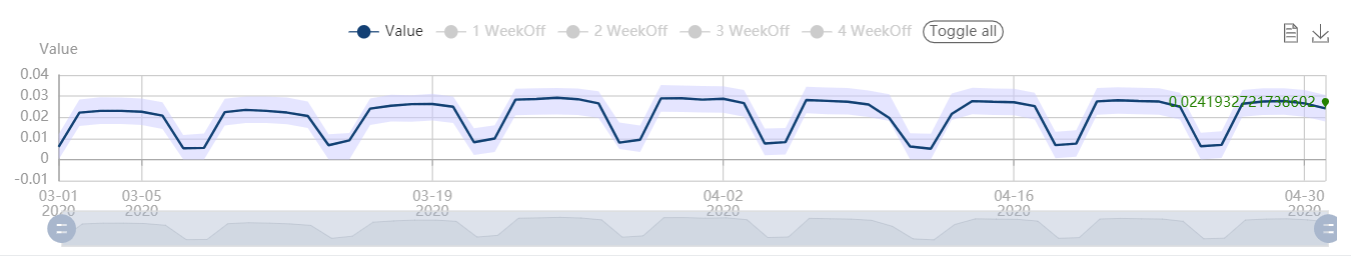}
	    \label{case1-b}}
    \subfigure[Detection results of Single Model]{
		\includegraphics[width=0.45\textwidth, height=3.5cm]{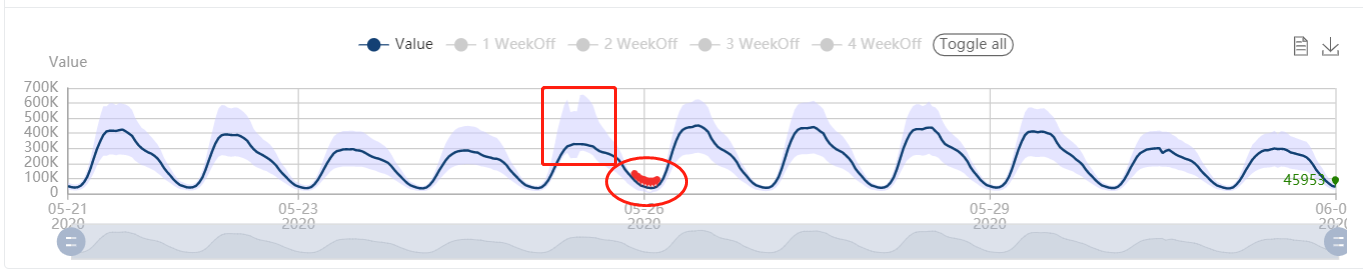}
		\label{case2-a}}
	 \subfigure[Detection results of Auto-Selector]{
	    \includegraphics[width=0.45\textwidth, height=3.5cm]{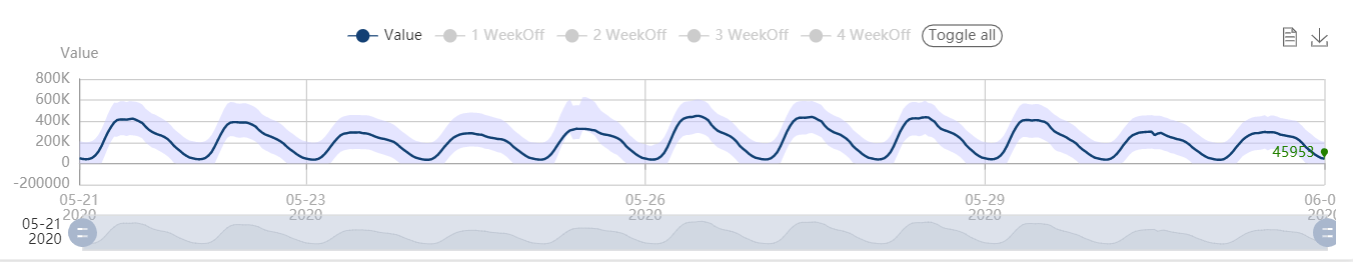}
	    \label{case2-b}}
	 \caption{Case Study}
	 \label{fig:case_study}
\end{figure*}
We study two cases in our service as shown in Figure~\ref{fig:case_study}. This first case in~\ref{case1-a} is a series with weekly pattern with one day as its interval. While the interval of the second series in~\ref{case2-a} is one hour with multiple seasonality, i.e. daily and weekly patterns. By default, seasonal series will use S-H-ESD~\cite{hochenbaum2017automatic} to detect anomalies, with anomaly ratio $0.15$ on the day frequency series and $0.18$ on the hour frequency. False alerts (red circles in the figures) are reported by customers. Root causes for these two dissatisfaction cases are 1) the S-H-ESD algorithm is sensitive to small changes in the circular part of the series but noises are common in such scenarios; 2) the anomaly ratio is relative high in these two cases; 3) the boundaries on these false positive timestamps are too narrow compared with other timestamps whose values are larger (red rectangles in the figures). These two dissatisfaction cases can be resolved with \textbf{Auto-Selector} as it automatically selects the most suitable models for each time series. It recommends HBOS~\cite{goldstein2012histogram} with threshold $0.923$ for the first case (as shown in Figure~\ref{case1-b}) and SR~\cite{sr-cnn} algorithm with threshold $4.894$ for the second case (as shown in Figure~\ref{case2-b}). Moreover, the customized anomaly detection result tuning algorithm gives a better boundary to cover different value scales so that customers can obtain a suitable sensitivity with consistent tolerance on different timestamps.
\section{Related Works}
\label{related_work}
\subsection{Time-Series Anomaly Detection Model}
Time-Series anomaly detection models can be categorized into statistical, supervised and unsupervised approaches. Hypothesis testing \cite{rosner1983percentage}, wavelet analysis \cite{lu2008network}, SVD \cite{mahimkar2011rapid} and ARIMA \cite{zhang2005network} are the classic representatives in statistics literature. HBOS \cite{goldstein2012histogram} and S-H-ESD \cite{hochenbaum2017automatic} are the latest ones in this area. 

Although the traditional statistical methods are fast, their performance are not satisfactory in real applications. To this end, researchers have developed more advanced models. Opprentice \cite{liu2015opprentice} leverages statistical detectors as feature extractors to build a random forest classifier \cite{liaw2002classification} and outperforms other traditional detectors. With the inspiration from visual saliency detection, SR-CNN \cite{sr-cnn} borrows the Spectral Residual model\cite{hou2007saliency} from the visual saliency detection model. It outperforms current state-of-the-art methods by a large margin and especially improved F1-score by more than 20\% on Microsoft production data.

However, continuous labels are hardly achievable in industrial scenarios most of the time. DONUT \cite{xu2018unsupervised}, an unsupervised anomaly detection method based on Variational Auto-Encoder (VAE) \cite{doersch2016tutorial} is proposed on the other hand. It uses the reconstruction error to check if a point is anomaly or not.

\subsection{Ensemble Multiple Anomaly Detection Models}
There are studies on ensemble anomaly detection. RandNet \cite{chen2017outlier} uses a series of autoencoders as base detectors. In 2017, Google also leveraged deep learning neural networks (DNN, RNN, LSTM) \cite{shipmon2017time} to detect anomalies on their own datasets and achieved promising results.  

CARE \cite{rayana2016sequential} and SELECT \cite{rayana2016less} are two similar heuristic unsupervised learning methods. After obtaining the detection results for all the base detectors. They both takes multiple round of iterations to converge to a final result. These two methods have risks that the result from a poor detector will pollute the final result. In our solution, such kind of risks are avoided as only the most suitable detector is used to perform anomaly detection. Moreover, our solution is more efficient in time complexity as the detection result can come out in one pass rather than multiple iterations.

Yahoo EGADS \cite{yahooo2015EGADS} utilizes a collection of anomaly detection, change point detection and forecasting models with an anomaly filtering layer for scalable anomaly detection on timeseries data. It also implements two algorithms for threshold selection in its ``Alert'' layer based on (a) K$\sigma$ deviation and (b) density distribution.
\section{Conclusion}
Time-series anomaly detection is a critical module to ensure the quality of online service. An industrial anomaly detection system should be able to leverage the advantage of any single anomaly detection model. In this paper, we introduce an automated mechanism to select best anomaly detection model and its corresponding hyper-parameters. This automated mechanism has been used in our product and benefited the online detection. Its ability makes it easier to enable effective time-series anomaly detection and reduce the time-cost of improving unsatisfied detection cases as we can trigger re-selection automatically. In future, we will enhance the ability of \textbf{Parameter Estimator} and reduce the dependence of anomaly labels in building an accurate \textbf{Auto-Selector}. 

\bibliographystyle{ACM-Reference-Format}
\bibliography{sources}


\end{document}